\documentclass[runningheads]{llncs}
\usepackage{miccai_style}

\begin{document}

\maketitle              
\begin{abstract}
Foundation Vision-Language Models (VLMs) trained using large-scale open-domain images and text pairs have recently been adapted to develop Vision-Language Segmentation Models (VLSMs) that allow providing text prompts during inference to guide image segmentation.
If robust and powerful VLSMs can be built for medical images, it could aid medical professionals in many clinical tasks where they must spend substantial time delineating the target structure of interest.
VLSMs for medical images resort to fine-tuning base VLM or VLSM pretrained on open-domain natural image datasets due to fewer annotated medical image datasets; this fine-tuning is resource-consuming and expensive as it usually requires updating all or a significant fraction of the pretrained parameters.
Recently, lightweight blocks called adapters have been proposed in VLMs that keep the pretrained model frozen and only train adapters during fine-tuning, substantially reducing the computing resources required.
We introduce a novel adapter, \textit{VLSM-Adapter}, that can fine-tune pretrained vision-language segmentation models using transformer encoders.
Our experiments in widely used CLIP-based segmentation models show that with only $3$ million trainable parameters, the VLSM-Adapter outperforms state-of-the-art and is comparable to the upper bound end-to-end fine-tuning.    
The source code is available at: \url{https://github.com/naamiinepal/vlsm-adapter}.
\end{abstract}
\section{Introduction}
\label{sec:introduction}

The early 2010s saw the initial success of Deep Learning in single-domain tasks such as image classification or language translation when deep neural networks could learn powerful representation using large-scale images \cite{deng2009imagenet,he2016deep} or texts \cite{fellbaum2010wordnet}.
As openly available large-scale annotated data lacked medical images, transfer learning was widely used where networks are initialized using weights obtained from pretraining in natural images such as ImageNet \cite{deng2009imagenet} and are further fine-tuned in domain-specific smaller datasets \cite{yosinski2014transferable}. 
Recently introduced foundation Vision-Language Models (VLMs) can learn powerful joint representation from large-scale images-text pairs and can be adapted to a wide range of tasks, including dense prediction tasks to develop Vision-Language Segmentation Models (VLSMs), which allow providing text prompts during inference to guide image segmentation.
VLSMs are attractive in the medical domain because robust and powerful VLSMs could aid medical professionals in many clinical tasks requiring tedious and time-consuming delineation of the target structure of interest.

VLSMs have separate vision and language encoders or joint vision-language encoders followed by a decoder or a mask-generating network that is trained end-to-end (E2E) \cite{wang2022cris}, or separately using frozen encoder parameters obtained from VLM pretraining \cite{luddecke2022image}.
The most popular VLM, widely adapted to create different VLSMs, is Contrastive Language-Image Pretraining (CLIP) \cite{radford2021learning}, which uses separate vision and language encoders.
It learns joint vision-language representation by projecting images and texts into a shared embedding space through learnable parameters that bring semantically similar image-text pairs close while pushing dissimilar pairs further apart.
Various VLSMs \cite{luddecke2022image,wang2022cris,xu2023side} have leveraged this multimodal semantic information captured by CLIP to train a segmentation model for an open-vocabulary segmentation task.
Yu et al. \cite{yu2023zero} used a pretrained self-supervised mask proposal network and CLIP to realize the zero-shot referring image segmentation on the open domain without additional training.

Although open-domain VLMs show impressive zero-shot or few-shot performances in downstream tasks, adapting them to medical image segmentation requires further fine-tuning \cite{adhikari2023synthetic,poudel2023exploring}.
This fine-tuning usually requires updating all~\cite{howard2018universal} or a significant fraction (usually last layers) of the pretrained parameters~\cite{long2015learning}, which is expensive because VLMs are much larger than popular image-only models (a few to several hundred million parameters).
Different methods have been proposed to efficiently fine-tune these foundation models, often called Parameter Efficient Fine-Tuning (PEFT) techniques.
The two most popular PEFT techniques are LoRA \cite{hu2022lora} and Adapters \cite{houlsby2019parameter} --- both of them adjust the intermediate representations of the pretrained models often using lightweight networks parallel to the pretrained ones with only slight differences.
Since adapters have been explored more in vision-language settings \cite{gao2023clip,song2024meta} compared to LoRA, we focus on adapters as a method for PEFT VLSMs for the scope of this paper.

Adapters are small networks with much fewer parameters that can be plugged into existing pretrained architectures, and then only adapters are trained while keeping pretrained weights frozen during fine-tuning.
VL-Adapter \cite{sung2022vl} reused the pretrained VLMs for vision-text tasks like image captioning and visual questioning-answering.
Although a few methods have been proposed for VLM-based classification tasks, no adapters are studied for E2E-trained VLSMs for further fine-tuning.
Side-Adapter Network (SAN) \cite{xu2023side} introduced ViT \cite{dosovitskiy2020image} as an adapter network, parallel to the CLIP's encoders, that generates segmentation masks for image-text inputs.
This paper proposes learnable adapter networks to fine-tune already trained VLSMs, as \textit{VLSM-Adapter}, which adapts the intermediate learned representations for domain-specific datasets while preserving the already learned weights from large-scale pretraining.
We add learnable adapter modules to a variant of VLSM, CLIPSeg \cite{luddecke2022image}, introducing $3$ million trainable parameters, which perform on par with the same model's E2E fine-tuning despite having almost $50$ times fewer trainable parameters.

The main contributions of this paper are:

\begin{itemize}
    \item We introduce novel adapter modules to efficiently fine-tune pretrained VLSMs to domain-specific smaller datasets using only a few learnable parameters.
    \item Our experiments and results on medical datasets with diverse modalities indicate that fine-tuning only the adapter modules for small datasets is better than E2E fine-tuning for VLSMs.
    \item We provide an ablation study on the positioning of adapter modules and show that introducing adapters deeper into the intermediate representations of the pretrained models results in better performance.
\end{itemize}

\section{Adapters for CLIP-based VLSMs}
\label{sec:adapters}

\subsection{Problem Statement}
\label{sec:problem_statement}

An encoder-decoder architecture-based pretrained model for vision-language segmentation model is frozen, while adapter modules with a much smaller number of parameters compared to the original frozen network are introduced to fine-tuning in a smaller training set comprising of the triplets: $D=\{(v_i,l_i,m_i)\}_{i=1}^S$.
Here, $S$ is the number of training samples, $v_i$, $l_i$, and $m_i$ represent the image input, text prompt, and target mask of the $i^{th}$ data point, respectively. 
The input images are RGB images and targets are their corresponding binary masks, i.e., $v_i \in \mathbb{R}^{H \times W \times 3}$, and $m_i \in \{0,1\}^{H \times W}$, respectively.

\subsection{Adapter Formulation}
\label{sec:our_approach:adapter}

Adapter modules \cite{houlsby2019parameter} are the non-linear projection blocks that adapt the representations of the pretrained models to a downstream task without changing their parameters, enabling the representations learned by the pretrained models to be used for other tasks.
\cref{eq:adapter} represents the basic block of an adapter network.

\begin{equation}
\label{eq:adapter}
    f' = Adapter(f) = f + \sigma(\psi(f \cdot W_1) \cdot W_2)
\end{equation}

Here, $f$ is the representation learned by the pretrained model, $f'$ is the adapted features, and $W_1$ and $W_2$ are learnable adapter weights.
$\psi$ and $\sigma$ are non-linear activation functions, which, in most cases, are the same type.
The adapter weights are initialized as $W_1 \in \mathbb{R}^{d \times d'}$, $W_2 \in \mathbb{R}^{d' \times d}$, where $d' \leq d$. 
The size of the input tensor must not change while exiting the adapters because they have to be used by the subsequent pretrained layers, i.e., $\{f,f'\} \in \mathbb{R}^{\ldots \times d}$.

\subsection{Proposed VLSM-Adapter}
\label{sec:vlsm_adapter}

\begin{figure}[ht]
    \centering
    \includegraphics[width=0.88\textwidth]{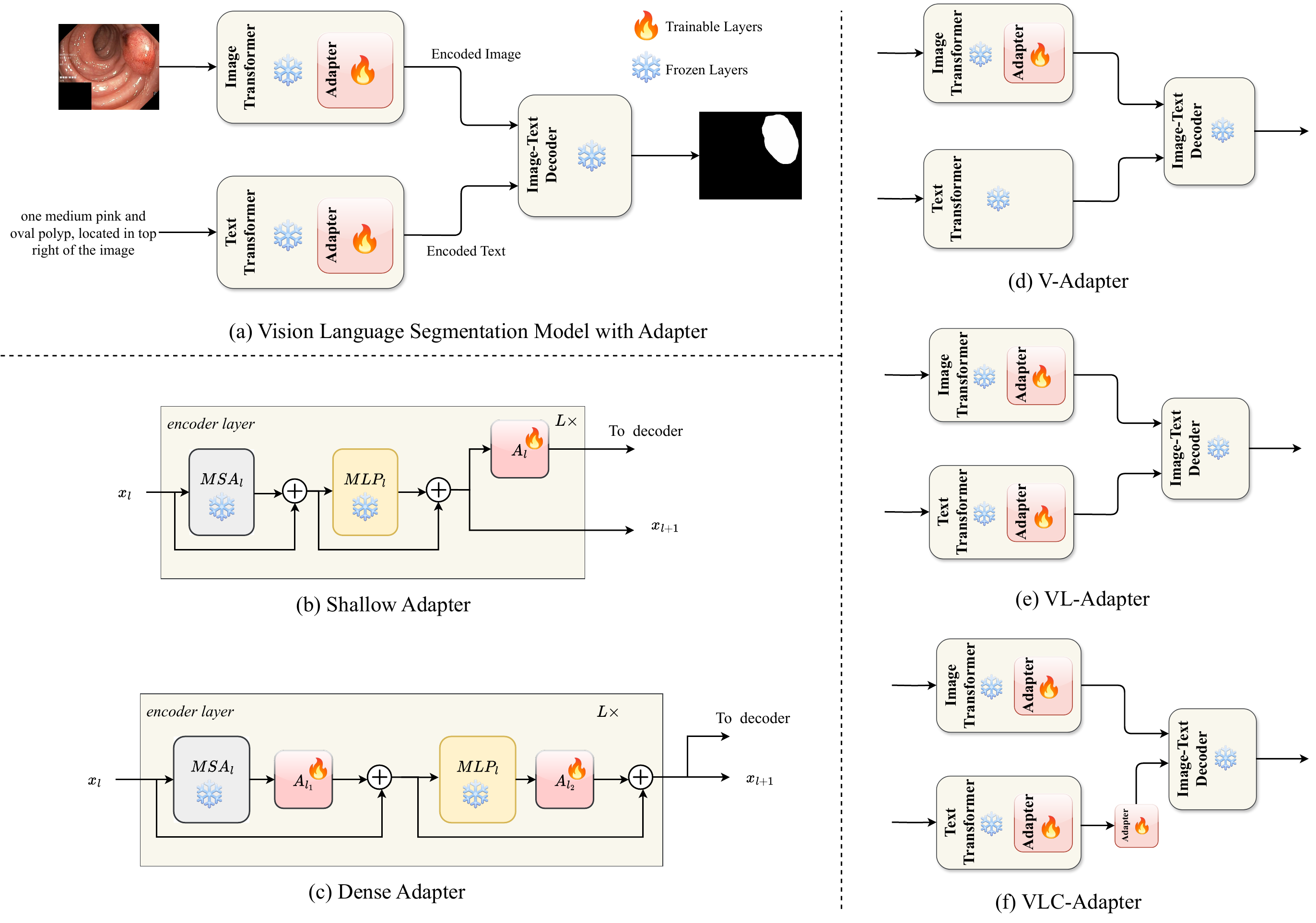}
    \caption{
        \textbf{Overall architecture of the proposed VLSM-Adapter module.}
        $MSA_l$, $MLP_l$, and $A_{l_*}$ stand for multi-head self-attention block, multi-layer perceptron, and adapter layers, respectively, for the $l^{th}$ transformer layer. 
        \textbf{(a)} Adapters are connected to each transformer block in the text and image encoders.
        \textbf{(b)} The shallow adapter adds learnable layers to each transformer block output.
        \textbf{(c)} The dense adapter employs the learnable layers before each residual addition in each transformer block. 
        \textbf{(d,e,f)} Adapters have been configured with different positioning for text and image encoders.
    }
    \label{fig:architecture}
\end{figure}

As displayed in \cref{fig:architecture}, we introduce adapters to the encoder segments while keeping the decoder static to VLSM-Adapter.
The positional variation to introduce transformer block adapters provides three incremental VLSM-Adapter variants.
\textbf{(1) V-Adapter} has adapters in the image encoder layers.
\textbf{(2) VL-Adapter} adds adapters for text encoder layers.
\textbf{(3) VLC-Adapter} adds an extra layer to adapt text conditioning at the bottleneck layer.
Since CLIPSeg \cite{luddecke2022image} provides transformer encoders and a pretrained segmentation mask decoder, we have used it as a candidate model for our experiments to validate VLSM-Adapter.
Two variants of VLSM-Adapter in CLIPSeg networks have been implemented for fine-tuning: CLIPSeg \textit{Shallow Adapter (SA)} and CLIPSeg \textit{Dense Adapter (DA)}.

\subsubsection{CLIPSeg Shallow Adapter}
\label{sec:clipseg_sa}

The shallow adapter (SA) in CLIPSeg \cite{luddecke2022image} learns to project the pretrained encoder representations before feeding them to the decoder network (\cref{fig:architecture}b).
The adapters are introduced at the skip connections of the CLIPSeg's encoders used by the vision-language decoder to predict segmentation masks.
Since the original CLIPSeg model \cite{luddecke2022image} used skip connections from $L_t \in \{3,6,9\}$ transformer \cite{vaswani2017attention} layers in the image encoder, three adapter layers are introduced at these connections.
A similar strategy adds a skip connection with adapter modules for $L_t$ layers in the text encoder.
CLIPSeg SA introduces $d' = 512$ as the hidden dimension of the adapter block, resulting in $4.2$ million trainable parameters.

\subsubsection{CLIPSeg Dense Adapter}
\label{sec:clipseg_da}

The dense adapter (DA) in CLIPSeg learns to adjust the representation of the successive layers of the encoders before feeding to the decoder network (\cref{fig:architecture}c). 
Following Houlsby et al. \cite{houlsby2019parameter}, we apply adapters before the two residual connections in each attention layer; two adapter blocks are used in each self-attention layer.
We use adapter block up to $\max(L_t) = L_T = 9$ attention layers of the image encoder because, beyond the $L_T$ layer, the intermediate representations remain unused by the decoder.
Similarly, DA implements the same pattern of adapters for the text encoder. 
DA also uses an adapter in CLIPSeg's text conditioning embeddings \cite{luddecke2022image} to ensure consistency with SA.
The hidden dimension of the block is $d'=64$, which introduces only $3$ million trainable parameters.
The empirical results of \cref{tab:fine-tuning} exhibit that DA surpasses SA in performance despite having fewer parameters.

The principal difference between SA and DA is that DA adapts the activations of each encoder block before feeding to the next one.
In contrast, SA adapts the extracted internal activations fed to the decoder.

\section{Experiments}
\label{sec:experiments}

\subsection{Datasets}
\label{sec:datasets}

Recently, Poudel et al. \cite{poudel2023exploring} proposed a wide range of automatic prompt generation methods and benchmark fine-tuning of different CLIP-based VLSMs in eight medical imaging datasets from diverse modalities, including five non-radiology and three radiology datasets. 
Following the convention of that work, we use their text prompts and the same splits of datasets.
Their proposed method generated multiple text prompts for an image-mask pair; for our empirical analysis, we randomly sample a text prompt among many to generate an image-mask-text triplet while iterating through the datasets.

Among the non-radiology datasets, three of them are endoscopic images with the polyp segmentation task (Kvasir-SEG \cite{jha2020kvasir}, ClinicDB \cite{bernal2015wm}, and BKAI~\cite{ngoc2021neounet}), one with diabetic foot ulcer segmentation task (DFU \cite{kendrick2022translating}), and the last one has for skin-lesion segmentation (ISIC-16 \cite{gutman2016skin}).
Three different radiology images include segmentation of breast ultrasound (BUSI \cite{al2020dataset}), 2D-echocardiography (CAMUS \cite{leclerc2019deep}), and chest X-ray (CheXlocalize \cite{saporta2022benchmarking}).

\subsection{Baseline Methods}
\label{sec:baseline_methods}

We benchmark five models for our experimental analysis --- two (CLIPSeg \cite{luddecke2022image} and CRIS \cite{wang2022cris}) are trained with E2E fine-tuning, and three (SAN \cite{xu2023side}, CLIPSeg SA, and CLIPSeg DA) with adapter fine-tuning.
SAN can generate segmentation masks from image-text inputs by training a ViT block \cite{dosovitskiy2020image} along frozen CLIP~\cite{radford2021learning}.
We are the first to use adapters for the pretrained encoder-decoder model for vision-language segmentation tasks, such as CLIPSeg DA and CLIPSeg SA.
Since the adapter module proposed by Houlsby et al. \cite{houlsby2019parameter} is incompatible with convolutional encoders, they are not practiced with the CRIS~\cite{wang2022cris}.
We use dice-score (DSC (\%)), intersection-over-union (IoU (\%)), and Hausdorff distance at the $95^{th}$ percentile (HD95) as metrics --- all averaged over a dataset --- to evaluate the overall performance of the methods.  

\subsection{Implementation Details}
\label{implementation_details}

The training and inference of the baseline and proposed methods are performed in an NVIDIA RTX 3090.
We use floating-point-16 mixed-precision training with a batch size of $32$.
The initial learning rates for the DA and SA are $1e-3$ and $3e-4$, respectively, with a scheduler that scales them by a factor of $0.3$ if no decrease in validation loss is observed for $5$ consecutive epochs.
If no progress in the validation DSC (\%) is observed for the $20$ consecutive epochs, then the training is stopped; thus, there is no fixed number of training epochs.
The models are optimized with AdamW \cite{loshchilov2018decoupled} with a weight decay of $1e-3$.
Also, each experiment is subjected to three different seed values to test the consistencies of the methods and account for the randomness in sampling the prompts.
We combined dice and binary cross-entropy losses for the objective function, as shown by \cref{eq:loss_function}.

\begin{equation}
\label{eq:loss_function}
    \mathcal{L} = \lambda_d \cdot \mathcal{L}_{Dice} + \lambda_{ce} \cdot \mathcal{L}_{BCE}
\end{equation}

Here, $\lambda_d$ and $\lambda_{ce}$ are hyperparameters; we chose their values for our experiments as $\lambda_d=1.5$ and $\lambda_{ce}=1$.

\section{Results and Discussions}
\label{sec:results}

\subsubsection{Variants of VLSM-Adapter.}
\label{sec:variants_of_vlsm_adapter}

\begin{figure}[ht]
    \centering
    \includegraphics[width=\linewidth]{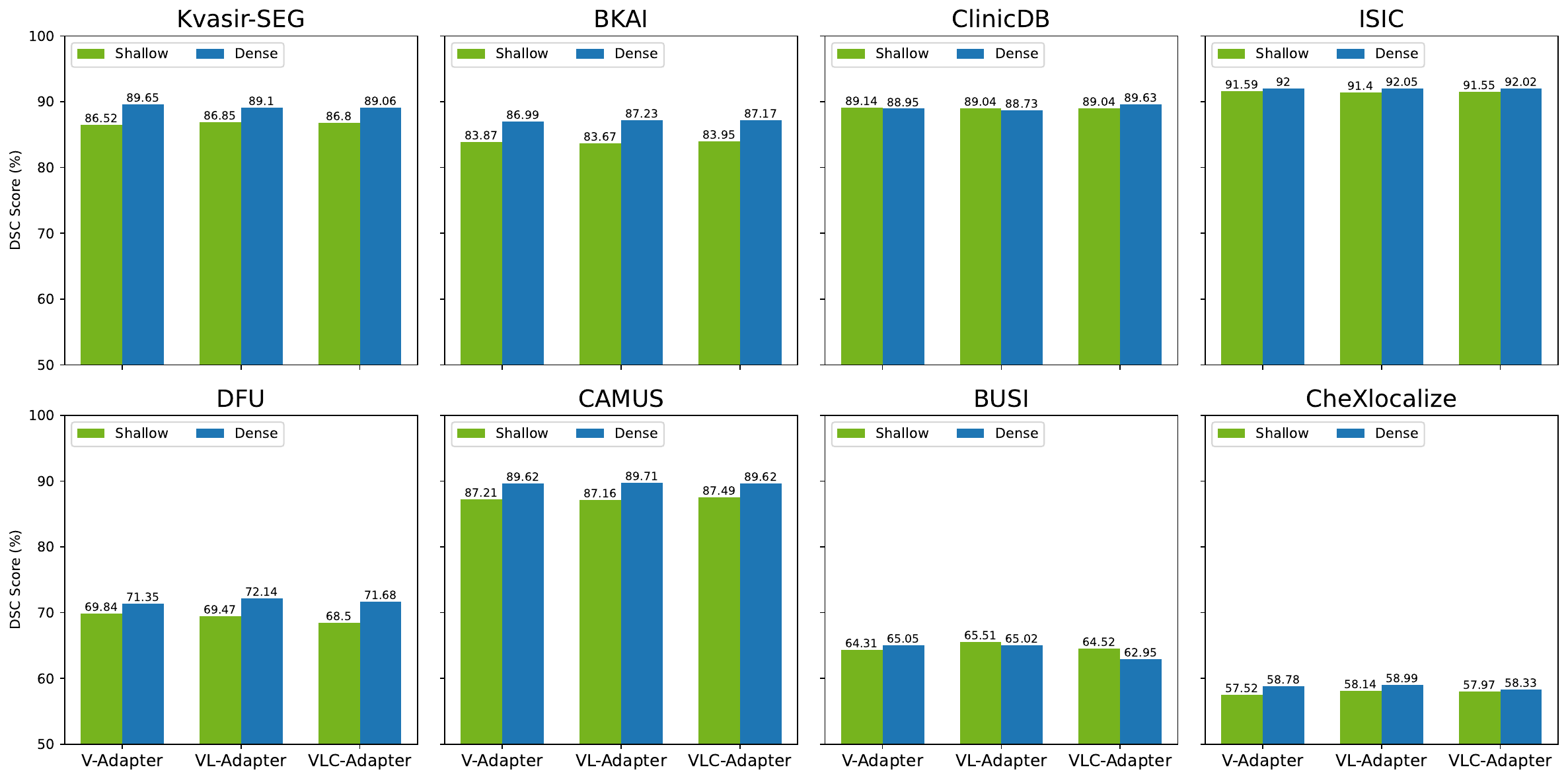}
    \caption{
    \textbf{Dice Score (\%) of variants of VLSM-Adapters across medical image datasets.}
    The dense adapter is better than the shallow adapter for almost all the datasets.
    }
    \label{fig:adapter-variants}
\end{figure}

In \cref{fig:adapter-variants}, we present the results of three different positioning of adapters in VLSMs as defined in \cref{sec:vlsm_adapter}.
The results show that VL-Adapter performs best in most of the datasets --- so, we have kept the performance of only this configuration in \cref{tab:fine-tuning}.
VLC-Adapter displays the optimal performance in the ClinicDB \cite{bernal2015wm} dataset.
V-Adapter exhibits the best score for Kvasir-SEG \cite{jha2020kvasir}, even superior to the upper bound set by CRIS \cite{wang2022cris} as indicated in \cref{tab:fine-tuning}.
Since the adapters are sensitive to their placements in encoder branches for generalizing domain-specific distribution of the datasets, one should evaluate different placements of adapters before selecting one variant. 
(see \cref{tab:adapter_position} in the supplementary section for more metrics)

\begin{table}[h]
    \centering
    \caption{
        \textbf{Evaluation of models across diverse medical image datasets. }
        $*M$ represents the number of trainable parameters in millions.
        \textbf{Bold} shows the best score among adapter fine-tuned models.
        \colorbox{Gray}{Gray} depicts the performance from E2E fine-tuning.
    }
    \label{tab:fine-tuning}
    \resizebox{\linewidth}{!}{%
    \begin{tabular}{l|r|| g g | r c c}
        \multirow{3}{*}{\textbf{Datasets}} & \multirow{3}{*}{\textbf{Metrics}} & \multicolumn{2}{c|}{\cellcolor{Gray} \textbf{Upper Bound}} &  \multicolumn{3}{c}{\textbf{Adapter Fine-tune}} \\
        & & CLIPSeg & CRIS & SAN &  CLIPSeg SA (\textbf{Ours}) & CLIPSeg DA (\textbf{Ours}) \\
        & & $150M$ & $147M$ & $8.4M$ & $4.2M$ & $\mathbf{3M}$ \\
        \hline
        \hline
        \multirow{3}{*}{Kvasir-SEG}  & DSC (\%) $\uparrow$ & $87.69$  & $89.43$ & $69.58$  & $86.85$ &  $\mathbf{89.10}$  \\
        & IoU (\%) $\uparrow$ & $81.72$ & $83.37$ & $58.05$ & $79.26$ & $\mathbf{82.39}$  \\
        & HD95 $\downarrow$ & $54.02$ & $55.23$ & $130.75$ & $52.18$ & $\mathbf{47.79}$\\
        \hline
        \multirow{3}{*}{BKAI} & DSC (\%) $\uparrow$ & $85.59$ & $92.62$ & $66.26$ & $83.67$ & $\mathbf{87.23}$\\
        & IoU (\%) $\uparrow$ & $77.52$ & $88.30$ & $54.58$ & $75.02$ & $\mathbf{79.81}$  \\
        & HD95 $\downarrow$ & $87.91$ & $49.80$ & $224.37$ & $87.79$ & $\mathbf{70.02}$\\
        \hline
        \multirow{3}{*}{ClinicDB}  & DSC (\%) $\uparrow$ & $88.58$ & $93.63$ & $81.36$ & $\mathbf{89.04}$ & $88.73$\\
        & IoU (\%) $\uparrow$ & $81.51$ & $88.74$ & $72.61$ & $\mathbf{81.93}$ & $81.84$ \\
        & HD95 $\downarrow$ & $19.30$ & $12.36$ & $38.42$ & $\mathbf{18.03}$ & $18.76$\\
        \hline
        \multirow{3}{*}{ISIC-16}  & DSC (\%) $\uparrow$ & $91.88$ & $91.49$ & $90.39$ & $91.40$ & $\mathbf{92.05}$\\
        & IoU (\%) $\uparrow$ & $85.76$ & $85.41$ & $83.61$ & $85.05$ & $\mathbf{85.98}$\\
        & HD95 $\downarrow$ & $60.93$ & $64.39$ & $87.25$ & $60.29$ & $\mathbf{54.38}$\\
        \hline
        \multirow{3}{*}{DFU} & DSC (\%) $\uparrow$ & $72.12$ & $74.01$ & $63.38$ & $69.47$ & $\mathbf{72.14}$\\
        & IoU (\%) $\uparrow$ & $61.61$ & $64.31$ & $51.63$ & $58.27$ & $\mathbf{61.42}$\\
        & HD95 $\downarrow$ & $38.24$ & $41.92$ & $60.10$ & $\mathbf{38.75}$ & $38.79$\\
        \hline
        \multirow{3}{*}{CAMUS} & DSC (\%) $\uparrow$ & $88.93$ & $91.29$ & $46.42$ & $87.16$ & $\mathbf{89.71}$\\
        & IoU (\%) $\uparrow$ & $80.69$ & $84.42$ & $31.81$ & $78.01$ & $\mathbf{81.85}$\\
        & HD95 $\downarrow$ & $16.69$ & $12.33$ & $175.81$ & $19.14$ & $\mathbf{14.16}$\\
        \hline
        \multirow{3}{*}{BUSI}  & DSC (\%) $\uparrow$ & $62.91$ & $67.50$ & $45.61$ & $\mathbf{65.51}$ & $65.02$\\
        & IoU (\%) $\uparrow$ & $55.52$ & $60.90$ & $35.27$ & $\mathbf{58.19}$ & $57.20$\\
        & HD95 $\downarrow$ & $72.98$ & $50.63$ & $152.10$ & $\mathbf{63.36}$ & $64.37$\\
        \hline
        \multirow{3}{*}{CheXlocalize}  & DSC (\%) $\uparrow$ & $58.51$ & $60.76$ & $44.37$ & $58.14$ & $\mathbf{58.99}$\\
        & IoU (\%) $\uparrow$ & $45.45$ & $47.99$ & $31.97$ & $44.84$ & $\mathbf{46.01}$ \\
        & HD95 $\downarrow$ & $537.57$ & $519.21$ & $724.55$ & $\mathbf{533.04}$ & $535.97$ \\
    \end{tabular}%
    }
\end{table}

\subsubsection{CLIPSeg Adapter outperforms E2E fine-tuning.}

With E2E fine-tuning for all radiology and non-radiology datasets, CLIPSeg-with-adapter shows superior performance for almost all the metrics compared to its counterpart, with no adapter module (see \cref{tab:fine-tuning}).
CLIPSeg with adapter modules, despite having $47$ times fewer trainable parameters than in an E2E setting, performing better than the latter shows the benefit of introducing learnable adapter modules with few trainable parameters in intermediate layers of CLIPSeg, rather than fine-tuning the whole model for small datasets.
Also, their performance is comparable to that of the state-of-the-art vision-only models for the individual datasets as reported by Poudel et al. \cite{poudel2023exploring}.

\subsubsection{Parameter-Metric Trade-off.}
In \cref{tab:fine-tuning}, the proposed CLIPSeg DA model performs better than the SAN \cite{xu2023side} model despite having $2.6$ times fewer learnable parameters. 
In the E2E fine-tuning cases, the CRIS \cite{wang2022cris} model has performed better than CLIPSeg \cite{luddecke2022image} in almost all the datasets.
The performance of CLIPSeg DA is on par with the CRIS model, even better in the ISIC-16 \cite{gutman2016skin} dataset, regardless of having $46$ times fewer parameters than CRIS.
Also, this drop in metrics may not be significant in some scenarios with a high computation constraint, which is precisely where our proposed adapter models shine.

\subsubsection{SA vs. DA.}

\begin{figure}[h]
    \centering
    \includegraphics[width=\linewidth]{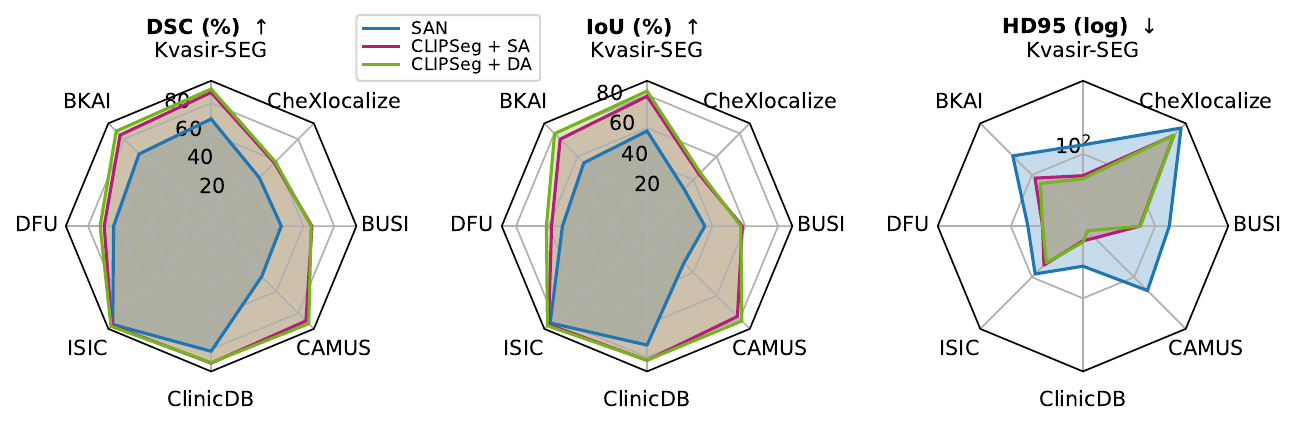}
    \caption{
    \textbf{Evaluation of Adapter-finetuned models.}
    Our methods perform better than SAN, with Dense Adapter (DA) performing the best across diverse datasets.
    }
    \label{fig:sa-da-san-spider}
\end{figure}

The DA network performs better than the SA network for most datasets, except for ClinicDB \cite{bernal2015wm} and BUSI \cite{al2020dataset}; even in those two datasets, the metrics of the DA network are on par with that of SA, as in \cref{tab:fine-tuning,fig:sa-da-san-spider}.
Since there are more adapter layers in DA, we suspect the layers can adjust the internal representations of the pretrained models more finely compared to SA.
Also, although the SA network has a broader adapter dimension of $512$, it cannot outperform the DA network, which has only a $64$ adapter dimension.
This signifies that deeper adapter networks can capture complex representations despite having smaller projection dimensions.

\section{Conclusion and Future Direction}
\label{sec:discussion_and_conclusion}

We present a VLSM-Adapter module that adjusts to the downstream segmentation tasks without changing the parameters of the pretrained encoder-decoder architecture.
We show that updating only the adapter parameters achieves on par performance to E2E fine-tuning, and is even better than the latter for some datasets.
The dense adapter variant performing better in most cases than the shallow adapter one, despite having fewer parameters, shows that tweaking the internal representations of the pretrained models finely in smaller dimensions --- dense adapter --- is more crucial than coarsely adapting the representations in a higher dimensional space --- shallow adapter.
Also, one should be open to experimenting with positioning the adapters in vision or text encoder branches.

In this work, we have only used a VLSM adapter for semantic segmentation in the medical domain.
The performance of adapter modules on other segmentation tasks for different domains is yet to be explored.
Additionally, this paper does not benchmark the performance of the models with adapters in the language-only branch because of the higher influence of the image encoder in the decoder of CLIPSeg; the decoder has skip connections from intermediate representations of the image encoder.
Future works can study the performance of models with adapters in the language encoder comparing it with the ones we demonstrated.

VLSM-Adapter also opens the pathways to continual learning and multi-task learning machines for VLSMs as specialized adapters could be trained for new data or tasks while keeping the core architecture frozen to prevent forgetting.
These adapters allow efficient fine-tuning of large pretrained VLSMs for medical image segmentation where there are often datasets with small sizes.

\begin{credits}
\subsubsection{\discintname}
The authors have no competing interests to declare that are
relevant to the content of this article.
\end{credits}

%
%
%
\bibliographystyle{splncs04}
\bibliography{references}

%
%
%
%
%
\appendix
\section*{Supplementary Material: VLSM-Adapter}
\label{appendix}





\begin{table}[h]
    \centering
    \caption{
        Evaluation of variants of VLSM-Adapters across medical image datasets.
        \textbf{Bold} shows the best score among all the models, and \underline{underline} represents the superior performance among shallow and dense adapters for the same setting.
    }
    \label{tab:adapter_position}
    \resizebox{\linewidth}{!}{%
    \begin{tabular}{l|r||cc|cc|cc}
    
        \multirow{3}{*}{\textbf{Datasets}} & \multirow{3}{*}{\textbf{Metrics}} & \multicolumn{6}{c}{\textbf{Adapter Configuration}}\\
        \cline{3-8}
        & & \multicolumn{2}{c|}{\textit{V-Adapter}} &  \multicolumn{2}{c|}{\textit{VL-Adapter}}& \multicolumn{2}{c}{\textit{VLC-Adapter}}\\
        \cline{3-8}
        & & Shallow& Dense & Shallow& Dense & Shallow&Dense \\
        \hline
        \hline
        \multirow{3}{*}{Kvasir-SEG} & DSC (\%) $\uparrow$ & $86.52$&  $\underline{\mathbf{89.65}}$& $86.85$ & $\underline{89.10}$   & $86.80$ & $\underline{89.06}$\\
        & IoU (\%) $\uparrow$ & $79.05$& $\underline{\mathbf{82.90}}$& $79.26$ & $\underline{82.39}$   & $79.33$ & $\underline{82.28}$\\
        & HD95 $\downarrow$ & $53.03$& $\underline{\mathbf{43.48}}$& $52.18$ & $\underline{47.79}$ & $53.18$ & $\underline{48.82}$\\
        \hline
        \multirow{3}{*}{BKAI} & DSC (\%) $\uparrow$ & $83.87$& $\underline{86.99}$& $83.67$ & $\underline{\mathbf{87.23}}$ & $83.95$ & $\underline{87.17}$\\
        & IoU (\%) $\uparrow$ & $75.35$& $\underline{79.60}$& $75.02$ & $\underline{\mathbf{79.81}}$   & $75.61$ & $\underline{79.74}$\\
        & HD95 $\downarrow$  & $80.73$& $\underline{\mathbf{64.48}}$& $87.79$ & $\underline{70.02}$ & $79.76$ & $\underline{64.72}$\\
        \hline
        \multirow{3}{*}{ClinicDB} & DSC (\%) $\uparrow$ & $\underline{89.14}$& $88.95$& $\underline{89.04}$ & $88.73$ & $89.04$ & $\underline{\mathbf{89.63}}$\\
        & IoU (\%) $\uparrow$  & $82.02$ & $\underline{82.19}$ & $\underline{81.93}$ & $81.84$  & $81.95$ & $\underline{\mathbf{82.89}}$\\
        & HD95 $\downarrow$ & $\underline{18.98}$ & $19.12$ & $\underline{18.03}$ & $18.76$ & $17.77$ & $\underline{\mathbf{17.09}}$\\
        \hline
        \multirow{3}{*}{ISIC-16} & DSC (\%) $\uparrow$ & $91.59$ & $\underline{92.00}$& $91.40$ & $\underline{\mathbf{92.05}}$ & $91.55$ & $\underline{92.02}$\\
        & IoU (\%) $\uparrow$ & $85.30$ & $\underline{85.96}$ & $85.05$ & $\underline{\mathbf{85.98}}$ & $85.22$ & $\underline{85.90}$\\
        & HD95 $\downarrow$ & $59.56$ & $\underline{52.46}$ & $60.29$ & $\underline{54.38}$ & $59.25$ & $\underline{\mathbf{51.76}}$\\
        \hline
        \multirow{3}{*}{DFU} & DSC (\%) $\uparrow$ & $69.84$ & $\underline{71.35}$ & $69.47$ & $\underline{\mathbf{72.14}}$ & $68.50$&$\underline{71.68}$\\
        & IoU (\%) $\uparrow$ & $58.57$& $\underline{60.59}$ & $58.27$ & $\underline{\mathbf{61.42}}$ & $57.33$ & $\underline{60.69}$\\
        & HD95 $\downarrow$ & $\underline{\mathbf{38.72}}$ & $40.17$ & $\underline{38.75}$ & $38.79$ & $\underline{40.07}$ & $41.12$\\
        \hline
        \multirow{3}{*}{CAMUS} & DSC (\%) $\uparrow$ & $87.21$ & $\underline{89.62}$ & $87.16$ & $\underline{\mathbf{89.71}}$ & $87.49$ & $\underline{89.62}$\\
        & IoU (\%) $\uparrow$ & $78.06$& $\underline{81.70}$ & $78.01$ & $\underline{\mathbf{81.85}}$ & $78.46$ & $\underline{81.71}$\\
        & HD95 $\downarrow$ & $18.93$& $\underline{14.18}$ & $19.14$ & $\underline{\mathbf{14.16}}$ & $18.21$ & $\underline{\mathbf{14.16}}$ \\
        \hline
        \multirow{3}{*}{BUSI} & DSC (\%) $\uparrow$ & $64.31$& $\underline{65.05}$& $\underline{\mathbf{65.51}}$ & $65.02$ & $\underline{64.52}$ & $62.95$\\
        & IoU (\%) $\uparrow$ & $56.82$& $\underline{57.18}$& $\underline{\mathbf{58.19}}$ & $57.20$ & $\underline{56.88}$ & $54.95$\\
        & HD95 $\downarrow$ & $69.23$& $\underline{\mathbf{59.41}}$& $\underline{63.36}$ & $64.37$ & $\underline{65.93}$ & $75.99$\\
        \hline
        \multirow{3}{*}{CheXlocalize} & DSC (\%) $\uparrow$ & $57.52$& $\underline{58.78}$& $58.14$ & $\underline{\mathbf{58.99}}$ & $57.97$ & $\underline{58.33}$\\
        & IoU (\%) $\uparrow$ & $44.36$ & $\underline{45.74}$ & $44.84$ & $\underline{\mathbf{46.01}}$  & $44.72$ & $\underline{45.33}$\\
        & HD95 $\downarrow$ & $\underline{535.44}$& $536.76$ & $\underline{\mathbf{533.04}}$ & $535.97$  & $542.80$ & $\underline{533.12}$\\
    \end{tabular}%
    }
\end{table}

\begin{figure}[h]
    \centering
    \includegraphics[width=\linewidth]{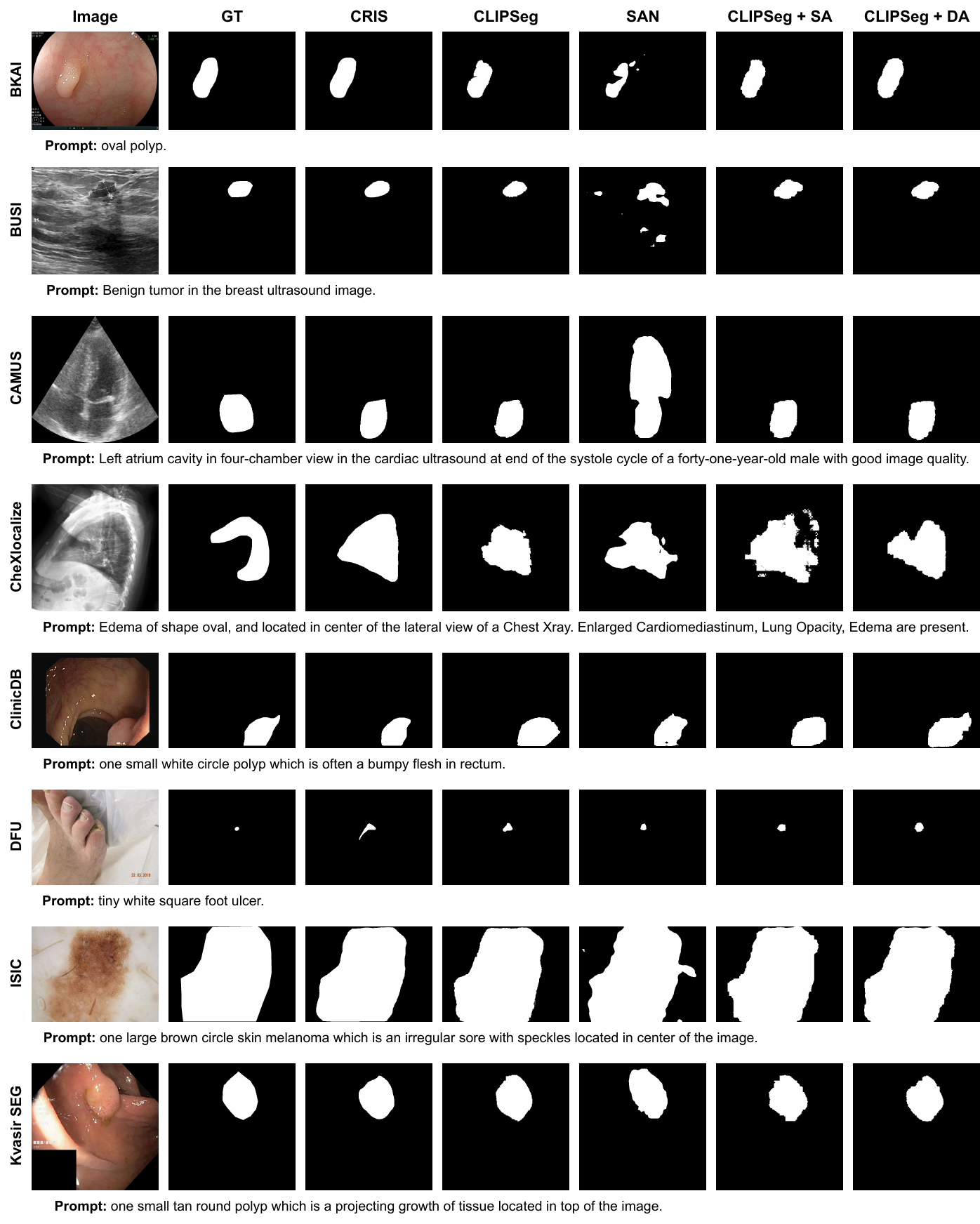}
    \caption{Qualitative results of different fine-tuned models.}
    \label{fig:qualitative-analysis}
\end{figure}

\end{document}